\pdfoutput=1

\documentclass[11pt]{article}

\usepackage[preprint]{acl}

\usepackage{times}
\usepackage{latexsym}

\usepackage[T1]{fontenc}

\usepackage[utf8]{inputenc}

\usepackage{microtype}

\usepackage{inconsolata}
\usepackage{amsmath}

\usepackage{graphicx}

\usepackage{xcolor}  
\usepackage{url}
\usepackage{booktabs} 
\usepackage{hyperref}
\hypersetup{
	colorlinks=true, 
    citecolor=blue, 
    linkcolor=blue,
    urlcolor=blue,
	pdfborder={0 0 0},
}
\usepackage{xcolor}
\usepackage{tcolorbox}
\usepackage{listings}
\usepackage{tcolorbox}
\usepackage{xcolor}
\usepackage{float}

\definecolor{commentcolor}{RGB}{0,100,0}   
\definecolor{keywordcolor}{RGB}{0,0,255}   
\definecolor{stringcolor}{RGB}{163,21,21}  
\definecolor{identifiercolor}{RGB}{0,0,0}  

\newtcolorbox{mybox}[1][]{
    title=#1,
    fonttitle=\small,
    fontupper=\small,
    left=2mm,
    right=2mm,
    top=1mm,
    bottom=0mm,
}

\lstdefinestyle{mystyle}{
    commentstyle=\color{commentcolor},
    keywordstyle=\color{keywordcolor}\bfseries,
    stringstyle=\color{stringcolor},
    identifierstyle=\color{identifiercolor},
    basicstyle=\ttfamily\lst@ifdisplaystyle\tiny\fi,
    breakatwhitespace=false,
    breaklines=true,
    captionpos=b,
    keepspaces=true,
    numbers=none,
    numbersep=5pt,
    showspaces=false,
    showstringspaces=false,
    showtabs=false,
    tabsize=2,
    xleftmargin=0pt,
}
\title{LaMDAgent: An Autonomous Framework for \\ Post-Training Pipeline Optimization via LLM Agents}

\author{
    Taro Yano \\
    NEC Corporation \\
    \texttt{taro\_yano@nec.com} 
    \\\And
    Yoichi Ishibashi \\
    NEC Corporation \\
    \texttt{yoichi-ishibashi@nec.com}
    \\\And
    Masafumi Oyamada \\
    NEC Corporation \\
    \texttt{oyamada@nec.com} \\
}

\begin{document}
\maketitle
\begin{abstract}
Large Language Models (LLMs) have demonstrated exceptional performance across a wide range of tasks. To further tailor LLMs to specific domains or applications, post-training techniques such as Supervised Fine-Tuning (SFT), Preference Learning, and model merging are commonly employed. While each of these methods has been extensively studied in isolation, the automated construction of complete post-training pipelines remains an underexplored area. Existing approaches typically rely on manual design or focus narrowly on optimizing individual components, such as data ordering or merging strategies. In this work, we introduce LaMDAgent (short for Language Model Developing Agent), a novel framework that autonomously constructs and optimizes full post-training pipelines through the use of LLM-based agents. LaMDAgent systematically explores diverse model generation techniques, datasets, and hyperparameter configurations, leveraging task-based feedback to discover high-performing pipelines with minimal human intervention. Our experiments show that LaMDAgent improves tool-use accuracy by 9.0 points while preserving instruction-following capabilities. Moreover, it uncovers effective post-training strategies that are often overlooked by conventional human-driven exploration. We further analyze the impact of data and model  size scaling to reduce computational costs on the exploration, finding that model size scalings introduces new challenges, whereas scaling data size enables cost-effective pipeline discovery.
\end{abstract}
\section{Introduction}
Large Language Models (LLMs) have undergone rapid development, demonstrating exceptional performance across diverse tasks and significantly impacting both academic and industrial domains, with the rise of high-performing proprietary models~\citep{DBLP:journals/corr/abs-2303-08774, claude35sonnet, gemini1_5} as well as open-sourced models~\citep{DBLP:journals/corr/abs-2407-21783, DBLP:journals/corr/abs-2412-15115, DBLP:journals/corr/abs-2412-19437, DBLP:journals/corr/abs-2404-14219}.
LLM development typically involves pre-training on large-scale web corpora followed by post-training with curated data~\citep{DBLP:conf/nips/Ouyang0JAWMZASR22}, with this study focusing on the latter stage due to the increasing emphasis on post-training driven by the release of models and datasets tailored for domain and task adaptation~\citep{DBLP:journals/corr/abs-2503-06072}.

In post-training, widely adopted approaches include Supervised Fine-Tuning (SFT) using human-created prompt-response pairs and Preference Learning based on preference labels for response pairs~\citep{DBLP:conf/nips/RafailovSMMEF23, DBLP:journals/corr/abs-2402-01306, DBLP:conf/aaai/00010LYHLW24, DBLP:conf/icml/MunosVCARGTGMFM24}. Furthermore, innovative techniques are rapidly evolving, such as autonomous training data generation and ``model merging'' that creates new models through arithmetic operations on different model parameters~\citep{DBLP:conf/icml/WortsmanIGRLMNF22, DBLP:conf/iclr/IlharcoRWSHF23, DBLP:conf/nips/YadavTCRB23}. To generate superior models, existing studies either manually build pipelines or focus on optimizing specific steps such as fine-tuning data orderings~\citep{DBLP:conf/nips/ChenRBWZSR23, DBLP:journals/corr/abs-2405-07490, DBLP:conf/emnlp/PattnaikMOYM24} or model merging strategies~\citep{DBLP:conf/naacl/IshibashiYO25, DBLP:journals/natmi/AkibaSTSH25}. However, full adaptation for target tasks requires combining these methods into integrated pipelines to optimize, yet automating this end-to-end process remains largely unexplored.

In this paper, we propose \underline{La}nguage \underline{M}odel \underline{D}eveloping Agent (LaMDAgent), a method that autonomously constructs post-training pipelines using LLM-based agents and continuously improves them based on feedback from the generated model's performance on target tasks. LaMDAgent treats heterogeneous model improving methods such as supervised fine-tuning, preference learning, or model merging in a unified manner and automates end-to-end post-training pipeline construction by exploring appropriate model generation methods, datasets, hyperparameters, and their optimal application order, thereby reducing the specialized knowledge and human costs required for pipeline construction.

Additionally, to reduce computational costs for LaMDAgent's exploration, we experimentally verify data size scaling and model size scaling, where data size scaling and model size scaling respectively involve exploring pipelines with smaller data quantities and model sizes, then transferring the discovered efficient pipelines to larger data quantities and model sizes. 

The contributions of this paper are as follows:
\begin{enumerate}
    \item We propose an LLM Agents-driven framework ``LaMDAgent'' that autonomously constructs and optimizes post-training pipeline. LaMDAgent treats heterogeneous model improving methods in a unified framework to optimize the entire pipeline in post-training, reducing the specialized knowledge and human costs required for pipeline construction.    
    \item In our experiments across two distinct settings, we show that LaMDAgent effectively improves mathematical capability by 3.7 points in average accuracy in Experiment 1 and enhances tool utilization accuracy by 9.0 points in Experiment 2 compared to strong baselines, while maintaining general capabilities through the discovery of novel pipelines that are not easily identified by humans.
    \item To reduce LaMDAgent’s exploration costs, we verify the effectiveness of data size scaling and model size scaling, finding that model size scalings introduces new challenges, whereas scaling data size enables cost-effective pipeline discovery.
\end{enumerate}

\section{Methodology}
\subsection{Overview}
We propose a novel method called \underline{La}nguage \underline{M}odel \underline{D}eveloping Agent (LaMDAgent) that  fully automates the construction and optimization of language model post-training pipelines using LLM Agents.
\begin{figure*}[thb]
    \centering
    \includegraphics[width=0.9\linewidth]{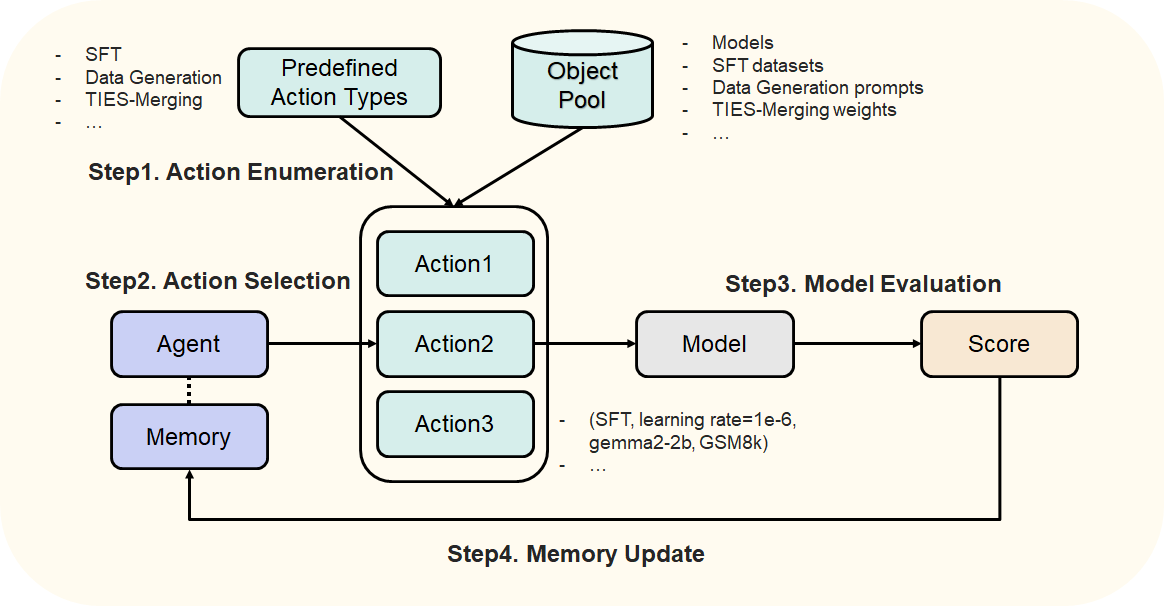}  
    \caption{Overview of our LaMDAgent framework. LaMDAgent first enumerates actions from predefined model improving action types and an object pool containing available data, models, parameters, and other objects (Step1. Action Eunumeration). Next, the agent selects an action based on memory acquired from previous trials and executes the selected action to generate a new model (Step2. Action Selection).
    Then evaluations on downstream tasks are conducted  (Step3. Model Evaluation). Based on the evaluation results of the newly generated model, the agent considers promising future directions and insights, updating the accumulated memory (Step4. Memory Update).}  
    \label{fig:method_overview}  
 \end{figure*}
Figure~\ref{fig:method_overview} illustrates the overview of our proposed method.
The proposed method aims to create better models by iteratively repeating the following four steps: 1. Action enumeration , 2. Action selection, 3. Model evaluation, and 4. Memory update. Details are described in the following sections.

\subsection{Action Enumeration }
For simplicity of explanation, we define the following terms:
\begin{itemize}
    \item Object: A concrete entity used in the model training pipeline, such as Llama 3 8B as a model or GSM8k as training data.
    \item Action: An action is a model improving method that takes multiple objects, including models, as input and outputs a new model. An action is defined by an action type such as "SFT" and the objects used, such as specific data, models, or hyperparameters. 
\end{itemize}
We obtain possible actions by enumerating all combinations of action types and objects.
Specifically, we use predefined action types and objects that include both pre-prepared datasets and models, as well as models and data obtained during the iteration.
For example, if we have the action type "SFT" is defined to take (base model, training data) as input objects, and we have Gemma2 2B as a base model and GSM8k and MATH as training data, then possible actions can be enumerated as (Gemma2 2B, GSM8k) and (Gemma2 2B, MATH).

\subsection{Action Selection}
We use the agent to select one promising model improvement action from possible actions. During action selection, we provide the agent with a prompt for action selection and parse its output to determine the action. 
In practice, rather than providing all action candidates and having the agent output a single action index, we first have the agent select an action type in one inference step and identify the required object types based on the action type. Then, we have the agent select objects in an another inference step to determine the final action.
We first decide on the action type to avoid action parsing failures that might occur if we give the agent the complex task of selecting the action type, understanding what object types are needed for each action type, and selecting objects without excess or deficiency. We select all objects in a single inference step rather than multiple steps to minimize order dependency in the selection process.

Action selection process can be written as
\begin{align}
    &a_{type} = Agent\left(g_{type}(m, l_{type})\right), \\
    &a_{obj} = Agent\left(g_{obj}(m, l_{obj}, a_{type})\right),
\end{align}
where $a_{type}$, $a_{obj}$, $g_{type}$ and $g_{obj}$ are determined action type, objects, prompt templates for selecting action types and objects, respectively. The prompts include memory $m$ summarizing experiences from past trials, candidate action types $l_{type}$, and objects $l_{obj}$. The actual $g_{type}$ and $g_{obj}$ used in our experiments is provided in Appendix~\ref{appendix:templates}. 
In preliminary trials, we observed mode collapse phenomena where the agent kept selecting the same action as steps progressed, so we explicitly included exploration directives in the prompt. 
We also added instructions to remove bias after observing that models generated at intermediate step $i$ named as "Model $i$" tended to be selected less frequently compared to initial models like "Model GSM8k".

\subsection{Model Evaluation}
We evaluate the selected actions based on the performance of the resulting model on target tasks and provide feedback to the agent through numerical scores. In single-task settings, the evaluation metric itself can be used as the score, but in multi-task settings, we need to aggregate metrics across tasks. To account for different scales of evaluation metrics across tasks, we define the multi-task score $s_{multi}$ using the following formula:
\begin{eqnarray}
    s_{multi} = \sum_{k} \alpha_k \cdot s_{single}^k,
\end{eqnarray}
where $s_{single}^k$ is the single-task reward for task $k$, and $0\leq \alpha_k \leq 1$ are scaling factors. In this research, we determine these factors so that the maximum contribution of each $s_{single}$ is uniform. For example, MT-Bench metrics range up to 10, while AceBench metrics reach 1, so we set the weights for each score as $\alpha_{MT} = 1/10, \alpha_{Ace}=1$.

\subsection{Memory Update}
\begin{figure*}[H]
  \centering
\vspace{-10pt}
\begin{mybox}
\vspace{-9pt}
\lstset{basicstyle=\scriptsize\ttfamily}
\begin{lstlisting}
You are a developer of Large Language Models (LLMs) that can improve models based on self reflections. You will be given results and memories of the previous improving trials. The results consist of actions and scores, where the scores are out of 1 point. And also, You will be provided with newly acquired trials. In a few sentences, update your memories based on the previous trials, memories, and new results.  

# Previous Results
<previous results>

# Previous Memories Acquired from Previous Trials
<previous memories>

# Newly acquired Results
<new results>

Updated Memory:
\end{lstlisting}
\vspace{-6pt}
\end{mybox}
\vspace{-9pt}
    \caption{Prompt template to update memory}
  \label{fig:eval template}
\end{figure*}
We update memories of the agent based on the feedback received for the selected action. 
A memory is a text summarizing experiences from the latest and past trials, and next promising directions to explore.
Memory at iteration $t$ is derived from the action at $t$-th iteration $(a_{type}^t, a_{obj}^t)$, score $s^t$, and template $g_{mem}$ as follows:
\begin{eqnarray}
    &&m^t = Agent(g_{mem}((a_{type}^t, a_{obj}^t, r^t), \nonumber \\ 
    &&\hspace{3mm}\{(a_{type}^{t^{\prime}}, a_{obj}^{t^{\prime}}, r^{t^{\prime}})\}_{t^{\prime} < t}, \{m^{t^{\prime}}\}_{t^{\prime} < t})).
\end{eqnarray}
The memory updating template $g_{mem}$ used in our experiments is provided in Appendix~\ref{appendix:templates}.

\section{Experiment 1: Teaching Multiple Skills to Base Models}
\begin{table*}[t]
\begin{center}
\caption{LaMDAgent effectively balances multiple skills and generalizes out-of-distribution: The main results of Experiment 1. LaMDAgent achieves the highest average performance (Avg) among compared methods. Notably, LaMDAgent Top-1 overperforms Fully Fine-Tuned by 3.7 points on math-related tasks (GSM8k, GSMSymbolic, NumGLUE1, NumGLUE2) on average, while maintaining performance on other tasks, demonstrating more effective multi-skill acquisition than simply mixing training data or merging specialist models.}
\scalebox{0.75}[0.75]{
  \begin{tabular}{llll|lllll|l}
    \toprule
    &\multicolumn{3}{c}{In-Distribution}&\multicolumn{4}{c}{Out-of-Distribution}&\\
    Method & GSM8k & CQA & TriviaQA & GSMSymbolic &NumGLUE1 &NumGLUE2 &SocialIQA &NQ & Avg \\
    \hline
    \multicolumn{5}{l}{\textbf{Baselines}} \\
    GSM8k-specialist & \textbf{0.320} & 0.001 & 0.000 &0.132 &\textbf{0.425} &\textbf{0.395} &0.030 &0.000 &0.163\\
    CQA-specialist & 0.018 & \underline{0.671} & 0.002 &0.007 &0.050 &0.034&\textbf{1.000} &0.008 &0.224\\
    TriviaQA-specialist & 0.046 & 0.027 & \textbf{0.675} &0.017 &0.050 &0.280 &0.905 &\textbf{0.269} &0.284\\
    TIES (Grid Search) & 0.105 & 0.559 & 0.562 &0.017 &0.175 &0.265 &0.999 &0.219 &0.363\\
    Fully Fine-Tuned & 0.254 & 0.622& \underline{0.672} &\underline{0.142} &0.325 &0.238 &\textbf{1.000} &\underline{0.256}&0.439 \\
    \midrule
    \multicolumn{5}{l}{\textbf{Proposed}} \\
    LaMDAgent Top-1 & \underline{0.284} & \underline{0.628} & 0.670 &\underline{0.145} &\underline{0.375} &\underline{0.302} &\textbf{1.000} &\underline{0.259} &\underline{0.458}\\
    LaMDAgent Top-2 & \underline{0.306} & 0.627 & \underline{0.673} &0.140 &\underline{0.350} &\underline{0.361} &\textbf{1.000} &0.248 &\textbf{0.463}\\
    LaMDAgent Top-3 & 0.267 & \textbf{0.674} & 0.658 &\textbf{0.146} &0.300 &0.256 &\textbf{1.000} &0.250 &\underline{0.444} \\
    \bottomrule
  \end{tabular}
}
\label{tab:main result1}
\end{center}
\end{table*}

\begin{table*}[htb]
\begin{center}
\caption{LLM-based action selection is effective: Ablation study results for LaMDAgent, with validation set scores shown in parentheses. Random action selection achieves only scores comparable to Fully Fine-Tuned, while LLM-based action selection achieves higher average performance. Additionally, the action space provided significantly impacts  generated model performances.}
\scalebox{0.8}[0.8]{
  \begin{tabular}{lllll}
    \toprule
    Method & GSM8k & CQA & TriviaQA & Avg \\
    \midrule
    Policy=LLM, Actions=(SFT, TIES)) & \textbf{0.284} (\textbf{0.350}) & \textbf{0.628} (\textbf{0.710}) & \textbf{0.670} (\textbf{0.750}) & \textbf{0.527} (\textbf{0.603}) \\
    Policy=Random, Actions=(SFT, TIES)) & 0.257 (0.280) & 0.594 (0.660) & 0.674 (0.730) & 0.508 (0.556) \\
    Policy=LLM, Actions=(TIES) & 0.032 (0.030) & 0.588 (0.670) & 0.575 (0.670) & 0.398 (0.456) \\
    \bottomrule
  \end{tabular}
}
\label{tab:ablation result1}
\end{center}
\end{table*}

\subsection{Experimental Setup}
We use Gemma2 2B~\citep{DBLP:journals/corr/abs-2408-00118}~\footnote{https://huggingface.co/google/gemma-2-2b} as our base model, and we target the following tasks in a multi-task setting: the arithmetic reasoning task GSM8k~\citep{DBLP:journals/corr/abs-2110-14168}, the commonsense reasoning task Commonsense QA (CQA)~\citep{DBLP:conf/naacl/TalmorHLB19}, and the reading comprehension task Trivia QA (TriviaQA)~\citep{DBLP:conf/acl/JoshiCWZ17}, all converted to 0-shot format. For out-of-distribution evaluation tasks, we use GSMSymbolic~\citep{DBLP:conf/iclr/MirzadehASTBF25}, which is a more complex version of GSM8k with rewritten numbers in the questions, generative arithmetic reasoning tasks from NumGLUE~\cite{DBLP:conf/acl/MishraMVSCBK22} Type1 (NumGLUE1) and Type2 (NumGLUE2), the social common sense reasoning task SocialIQA~\citep{DBLP:journals/corr/abs-1904-09728}, and the reading comprehension task Natural Questions (NQ)~\citep{DBLP:journals/tacl/KwiatkowskiPRCP19}.

While our approach can handle any action that produces a single model, the actions used in this experiment and their required objects are:
\begin{itemize}
    \item TIES-Merging (TIES): Model 1, Model 2, merge weight (fixed), merge density (fixed)
    \item Supervised Fine-Tuning (SFT): Model, SFT training data, learning rate (fixed)
\end{itemize}
TIES is a representative model merging technique, while SFT is a standard training method using log-likelihood maximization loss.
As initial objects, we prepared specialist models trained on 1,000 examples each from GSM8k, CQA, and TriviaQA using Gemma2 2B as the base model, hereafter referred to as GSM8k-specialist, CQA-specialist, and TriviaQA-specialist. We also use the training data same as specialist models along with an aggregated all data as initial objects for SFT. For hyperparameters, we fix merging weights of $(0.5, 0.5)$, density of 0.5, and learning rate for SFT as $1e-6$.
We use 100 examples from a different split as validation data, and test data was held out from both training and validation data. To eliminate variability from randomness, temperature was set to 0 during both pipeline exploration and testing.
For the agent LLM, we use gpt-4o-2024-08-06 and performed 100 iterations of action selection and feedback.

For details of evaluation methods, GSM8k involves parsing the final numeric answer from the prediction and exact matching with the ground truth, CQA requires the answer choice to be the form of "[[{choice}]]" and checking if the parsed value is correct, and TriviaQA uses exact match between normalized predictions and ground truth. 
For out-of-distribution tasks, GSMSymbolic, NumGLUE1, and NumGLUE2 uses the same evaluation method as GSM8k, SocialIQA uses the same as CQA, and NQ uses the same as TriviaQA.

For compared methods, in addition to the GSM8k, CQA, and TriviaQA specialists, we use TIES (Grid Search), which optimizes the weights of the three specialists through grid search, and Fully Fine-Tuned, which is trained on all available training data.
To evaluate the effectiveness of LLM-based action selection, we also compare with Policy=Random, Actions=(SFT, TIES)) which randomly selects actions for 100 iterations, and Policy=LLM, Actions=(TIES) which removes the SFT from predefined action types.

\subsection{Results}
The experimental results are shown in Table~\ref{tab:main result1}. LaMDAgent Top-$i$ refers to the model with the $i$-th highest accuracy on validation set among those generated by LaMDAgent. Bold values indicate the best performance among comparison methods, and underlined values indicate the top three.

\textbf{LaMDAgent outperforms baselines, enhancing math skills while preserving others.} Compared to the best baseline, Fully Fine-Tuned, LaMDAgent Top-1 shows 1.9 point improvement in overall accuracy (Avg) on the test set, demonstrating the effectiveness of the discovered pipeline. The improvement is particularly notable in arithmetic reasoning tasks, with LaMDAgent Top-1 overperforms Fully Fine-Tuned by 3.7 points on math-related tasks (GSM8k, GSMSymbolic, NumGLUE1, NumGLUE2) on average, while maintaining comparable performances on other tasks. These results suggest that, even with identical training data, appropriately combining model merging and training sequences using LaMDAgent can incorporate multiple skills more effectively than simple SFT on all the data.

\textbf{Training is more effective than model merging for acquiring multiple skills.} Interestingly, unlike findings in some previous works~\citep{DBLP:conf/emnlp/MorrisonSHKDD24, DBLP:journals/corr/abs-2410-14735}, in our experimental setting, Fully Fine-Tuned, which was trained on all data, outperformed TIES (Grid Search), which optimizes model merging weights, on all in-distribution tasks and 4 out of 5 out-of-distribution tasks, showing a 7.6 point higher average accuracy.

\textbf{Agent-based action selection is effective.} The ablation results in Table~\ref{tab:ablation result1} show that random action selection (Policy=Random, Actions=(SFT, TIES)) resulted in a 4.7 point decrease on the validation set and a 1.9 point decrease on the test set, demonstrating the effectiveness of agent-based action selection. 
This is likely because as iterations progress, random action selection tends to prioritize exploring combinations of model merging, which is less effective in this setting as shown in TIES results, over exploration of training data curricula. This occurs because as the number of models increases with iterations, the number of model merging action candidates grows quadratically, while the number of training candidates grows linearly, making the former more likely to be selected randomly.

\textbf{The choice of action space significantly impacts performance.} Removing SFT from the action space (Policy=LLM, Actions=(TIES)) led to decreases of 14.7 and 12.9 points on validation and test sets respectively, showing that the pre-defined action space significantly affects the final achievable accuracy.

\begin{figure}[thb]  
\includegraphics[width=\linewidth]{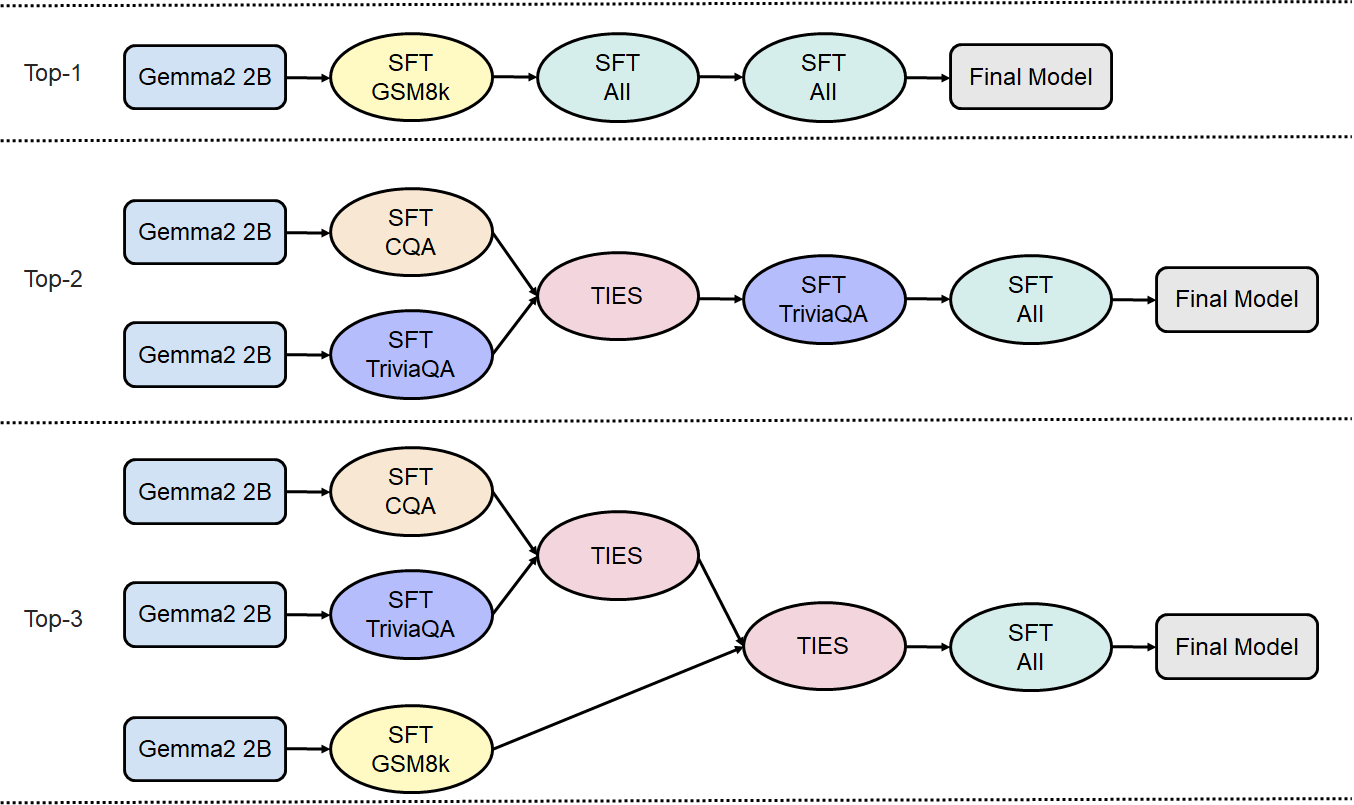}
    \caption{Top-1, Top-2, and Top-3 pipelines discovered in experiment 1.}
    \label{fig:top_pipelines_experiment1}  
 \end{figure}

\textbf{Discovered pipelines.} The highest-performing pipelines discovered by LaMDAgent are shown in Figure~\ref{fig:top_pipelines_experiment1}. The Top-1, 2, and 3 pipelines all have in common that they end with training on all data.
For Top-1, the result is consistent with findings~\citep{DBLP:conf/acl/DongYLLXL00ZZ24} that learning mathematical skills first before mixing with general skill data is beneficial for balancing mathematical skills like GSM8k with general skills like CQA and TriviaQA.
Interestingly, while model merging is typically used as a final refinement stage after training, in our experiments, pipelines that merge before training (Top-2 and Top-3) also performs well.

\section{Experiment 2: Enhancing Tool Usage Skills in Instruction-tuned Models}\label{sec:experiment 2}

\begin{figure}[thb]  
\includegraphics[width=\linewidth]{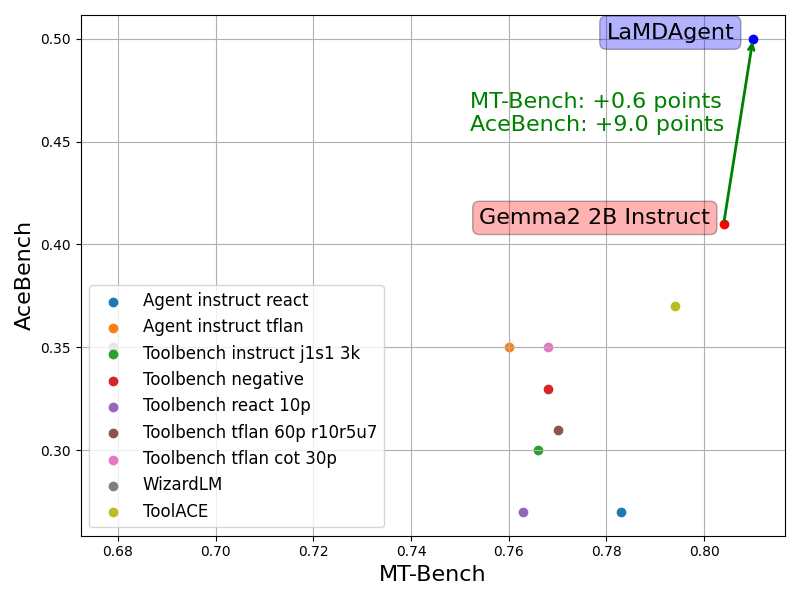}
    \caption{LaMDAgent significantly improves tool usage capability while maintaining instruction-following performance: The overall performance evaluation results of Experiment 2 indicate that LaMDAgent improves AceBench accuracy by 9.0 points while preserving the MT-Bench score. In contrast, naive fine-tuning approaches on either individual or full SFT datasets fail to enhance tool usage capabilities, suggesting that the task cannot be effectively addressed with such straightforward methods.}
    \label{fig: exp2_main}  
 \end{figure}

\begin{figure}[thb]  
\includegraphics[width=\linewidth]{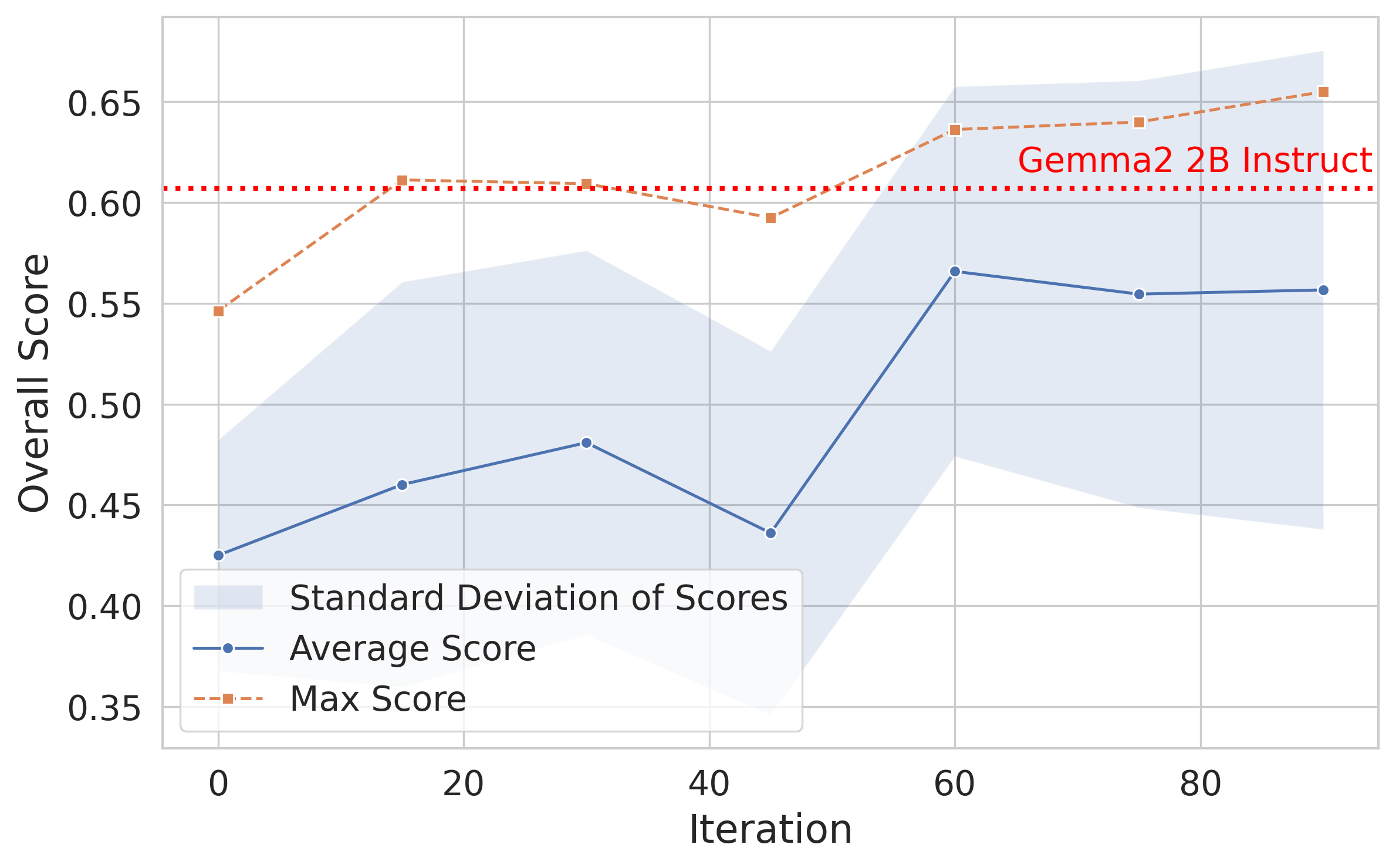}
    \caption{LaMDAgent learns from feedback to exploit promising actions while exploring unseen pipelines: The graph shows the Average Score, Max Score, and Standard Deviation recorded every 15 iterations. The consistent increase in average score indicates that the agent continues to learn from past feedback to exploit promising actions. The non-zero standard deviation through all iterations and improving max score implies that the agent maintains exploration to discover further improvement opportunities alongside exploitation.}
    \label{fig: exp2_stepwise_results}  
 \end{figure}
\subsection{Experimental Setup}
In a more realistic setting, we test whether LaMDAgent can enhance a specific skill (tool usage in this case) while maintaining the original instruction-following capabilities of a publicly available instruction-tuned model, Gemma2 2B Instruct~\footnote{https://huggingface.co/google/gemma-2-2b-it}. We use AceBench~\citep{chen2025acebenchwinsmatchpoint} to evaluate tool usage capabilities and the first turn of MT-Bench~\citep{DBLP:conf/nips/ZhengC00WZL0LXZ23} to evaluate instruction-following capabilities.
For action types, we adopt TIES and SFT as in Experiment 1. Initial objects include Gemma2 2B Instruct as the model, and for tool usage training data we use Agent-FLAN~\footnote{https://huggingface.co/datasets/internlm/Agent-FLAN}~\citep{DBLP:conf/acl/ChenLWZLLCZ24}, which includes Toolbench react10p, Toolbench tflan 60p r10r5u7, Toolbench tflan cot 30p, Agent instruct react, Agent instruct tflan, Toolbench instruct j1s1 3k, and Toolbench negative), ToolACE~\footnote{https://huggingface.co/datasets/Team-ACE/ToolACE}~\citep{DBLP:journals/corr/abs-2409-00920}, and general instruction-following data of WizardLM~\footnote{https://huggingface.co/datasets/\\WizardLMTeam/WizardLM\_evol\_instruct\_V2\_196k}~\citep{DBLP:conf/iclr/XuSZG0FTLJ24}. We randomly selected up to 1,000 examples from each of the 7 Agent-FLAN subsets, ToolACE, and WizardLM. As in Experiment 1, we use gpt-4o-2024-08-06 as the LLM for action selection and performed 100 iterations.

For evaluation, we use only turn 1 of MT-Bench for faster and more cost-effective assessment, with gpt-4o-2024-08-06 as the evaluator. For ACEBench, we report the accuracy in the Normal setting, which measures single-turn function call performance. The temperature parameter is set to 0 during both pipeline exploration and testing to eliminate randomness.

Compared methods include the Gemma2 2B Instruct, Individually fine-tuned models trained separately on each of the 9 training datasets, and a Fully fine-tuned model trained on all data. All fine-tuned models use the same hyperparameters as the SFT in LaMDAgent.

\subsection{Results} The overall performance evaluation results of Experiment 2 are summarized in Figure~\ref{fig: exp2_main}.
Also, figure~\ref{fig: exp2_stepwise_results} plots the average score, maximum score, and standard deviation of scores for models created by LaMDAgent at 15-iteration intervals.

\textbf{LaMDAgent enhances tool usage capabilities of Gemma2 2B Instruct while preserving instruction-following capabilities.} The best model generated by LaMDAgent achieves an MT-Bench score of 0.810, comparable to Gemma2 2B Instruct (0.804), while improving AceBench accuracy from 0.410 to 0.500—a 9.0 point improvement. This demonstrates successful enhancement of tool usage capabilities while maintaining instruction-following performance. In contrast, the all fine-tuned models, significantly degrades both instruction-following and tool usage capabilities.
A possible explanation for this: Gemma2 2B Instruct may have already paid an "alignment tax"~\citep{DBLP:conf/nips/Ouyang0JAWMZASR22} through extensive instruction tuning, and so unstable that additional tool-focused training could cause catastrophic forgetting of pre-training knowledge easily unless the training pipeline is carefully selected.
To support this hypothesis, as shown in Figure~\ref{fig: exp2_stepwise_results}, while LaMDAgent occasionally takes destructive actions, it learns to avoid them over time through feedback from downstream task, allowing the agent to automatically avoid such actions regardless of the cause.

\textbf{Exploiting from score-based feedback while exploring unseen pipelines.} As shown in Figure~\ref{fig: exp2_stepwise_results}, the average score continues to improve with iterations, confirming that the LaMDAgent framework effectively updates its memory to exploit promising actions. The continuous improvement in maximum score and non-zero values of standard deviation of scores suggests that the agent maintains exploration alongside exploitation.

\textbf{LaMDAgent is more effective when training and target distributions do not match.} The score difference between Fully Fine-Tuned and LaMDAgent in Experiment 1 was smaller than in Experiment 2, indicating that LaMDAgent provides greater benefits in Experiment 2.
This is because when trainining and target distributions are the same, simply minimizing the loss function on target tasks can yield good performances, whereas when distributions differ (as in Experiment 2), minimizing loss on all training data doesn't necessarily minimize loss on target data. Such scenarios represent effective applications for LaMDAgent.

\textbf{Discovered pipelines.} 
\begin{figure}[htb]  
\includegraphics[width=\linewidth]{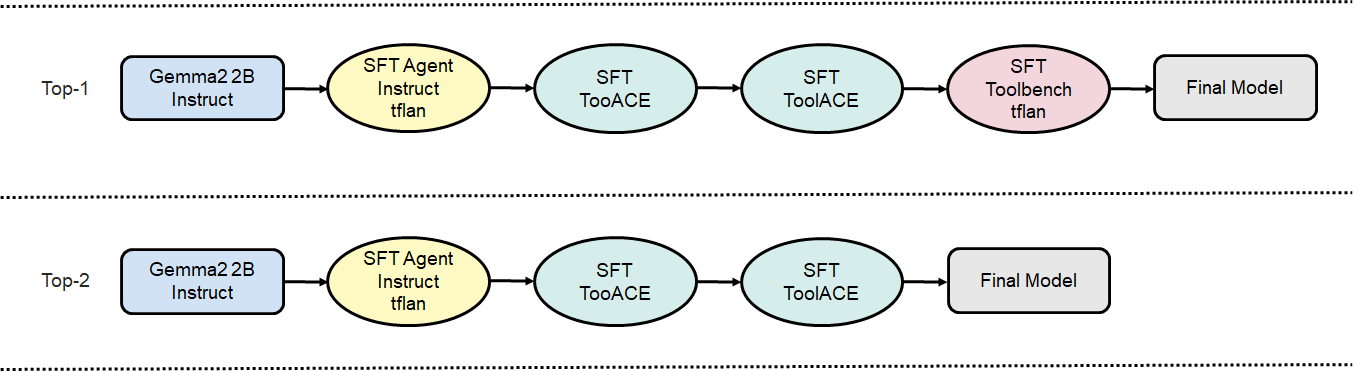}
    \caption{Top-1 and Top-2 pipelines discovered in experiment 2.}
    \label{fig:top_pipelines_experiment2}  
 \end{figure}
 Figure~\ref{fig:top_pipelines_experiment2} shows the Top-1 and Top-2 pipelines with the highest scores. Both pipelines train on Agent instruct tflan followed by ToolACE, suggesting these datasets were effective for AceBench. However, since the Fully Fine-Tuned model (which included these datasets) performs worse than the baseline Gemma2 2b Instruct, suggesting that excluding unnecessary data and establishing an appropriate training orderings are crucial. The Top-1 model's score evolution is 0.442 (SFT on Agent instruct tflan) → 0.592 (SFT on ToolACE) → 0.625 (SFT on ToolACE) → 0.655 (SFT on Toolbench tflan 60-r10r5u7), showing that similar performance improvements at each step cumulatively contributed to the final score, which cannot be easily identified by humans.

\section{Reducing Computational Cost}
In this section, we investigate the effectiveness of data size scaling and model size scaling inspired by pre-training scaling laws~\citep{DBLP:journals/corr/abs-2408-00118} as methods to reduce the computational cost of LaMDAgent's pipeline exploration.

\textbf{Data size scaling is effective.} Data size scaling is based on the expectation that pipelines with high scores on small data sizes will maintain high scores when data size is increased. This approach involves exploring effective pipelines with small data sizes, then scaling up the data within those pipelines.
For data size scaling to be effective, pipelines that outperform others with small data sizes must continue to outperform when data sizes are increased.

\begin{figure}[htb]  
\includegraphics[width=\linewidth]{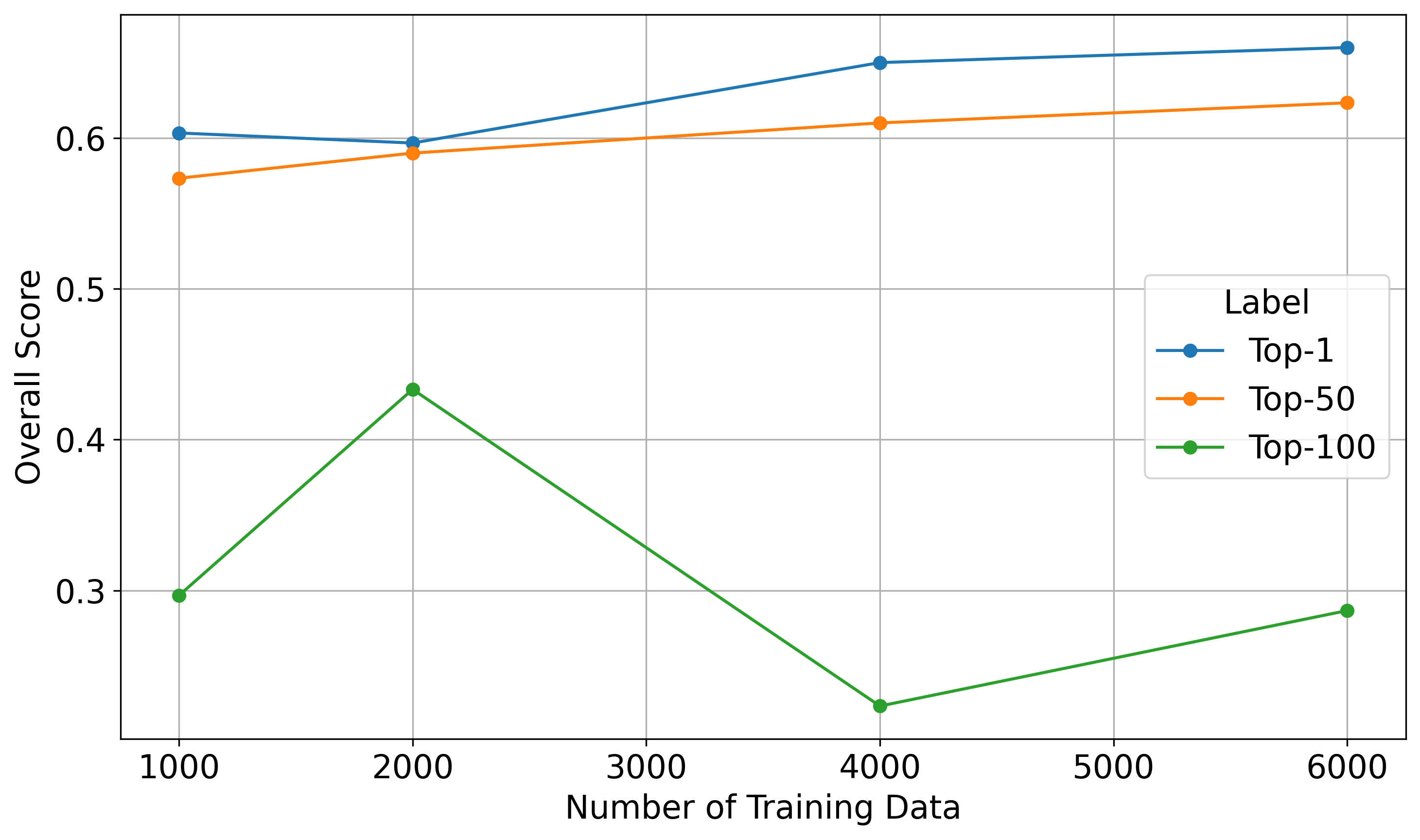}
    \caption{Data size scaling is effective for computational cost reduction: Overall score when scaling number of training examples in Top-k pipelines. The Top-1 pipeline consistently performs best, suggesting that effective pipelines at small scales remain effective with more data.}
    \label{fig: exp1_data_size_transfer}  
 \end{figure}

To verify the effectiveness of data size scaling, we examine how scores change when increasing the data in pipelines discovered in Experiment 1 by factors of 2, 4, and 6 times the exploration size. The results are shown in Figure~\ref{fig: exp1_data_size_transfer}. The Top-1 pipeline maintains the highest accuracy across all data sizes, demonstrating that pipelines with high accuracy on small data sizes maintain their advantage when scaled up, confirming the effectiveness of data size scaling for computational cost reduction. 

\begin{table}[h]
\begin{center}
\caption{Challenges exist in computational cost reduction via model size scaling: Evaluation scores of transferred pipelines on Gemma2 9B, suggesting that although some performance gaps are maintained, they sometimes diminish with model size scaling.}
\scalebox{0.8}[0.8]{
  \begin{tabular}{llllll}
    \toprule
    Method&Top-1 &Top-50 &Top-80 &Top-90 &Top-100\\
    \hline
     2B-based&0.603  &0.573 &0.553 &0.546 &0.297\\   
     9B-based&0.797 &0.803 &0.783 &0.783 &0.200  \\    
    \bottomrule
  \end{tabular}
}
\label{tab:model size transfer}
\end{center}
\end{table}

\textbf{Model size scaling has limitations.} Model size scaling is the model size version of scaling, which involves exploring effective pipelines with small models, then scaling up the models within those pipelines.

To verify the effectiveness of model size scaling, we change the base model from Gemma2 2B to Gemma2 9B to transfer the discovered pipelines in experiment 1. The results are shown in Table~\ref{tab:model size transfer}. When comparing Top-1 with Top-80, 90, and 100, which had score differences of more than 5 points with the 2B model, the Top-1 pipeline still achieves higher scores with 9B, showing that discovered pipelines remain effective when base model size increases. However, the score difference between Top-1 and Top-50, which was about 3 points with 2B models, reverses when scaling to the larger model, suggesting that small score gaps may disappear when increasing model size.
Therefore, in practice, rather than pursuing small pipeline differences that risk disappearing, diversifying action space to explore large score gaps may be an effective use case for LaMDAgent when expecting model size scaling.
\section{Related Work}

\textbf{LLM Agents.}
LLMs have evolved beyond chatbots to agents capable of executing diverse actions \citep{DBLP:journals/fcsc/WangMFZYZCTCLZWW24, DBLP:journals/chinaf/XiCGHDHZWJZZFWXZWJZLYDW25}. ReAct \citep{DBLP:conf/iclr/YaoZYDSN023} enables iterative reasoning via thought-action-observation loops, while Reflexion \citep{DBLP:conf/nips/ShinnCGNY23} introduces verbal learning from feedback on past trajectories. Key areas include web automation \citep{DBLP:journals/corr/abs-2503-23350, DBLP:conf/iclr/ZhouX0ZLSCOBF0N24, DBLP:conf/nips/DengGZCSWSS23} and tool use \citep{DBLP:conf/nips/PatilZ0G24, DBLP:journals/fcsc/QuDWCWYXW25, DBLP:conf/iclr/QinLYZYLLCTQZHT24}.  To our knowledge, this is the first work to automate post-training using LLM agents, treating improvement strategies as actions and model scores as rewards.

\textbf{Curriculum Learning in Post-Training.}
Post-training performance is sensitive to the order of training data. SKILL-IT \citep{DBLP:conf/nips/ChenRBWZSR23} prioritizes samples effective on validation sets. DMT \citep{DBLP:conf/acl/DongYLLXL00ZZ24} uses a two-stage process starting from specialized to general tasks. \citet{DBLP:journals/corr/abs-2405-07490} proposes reordering based on attention scores, query length, and loss, while Curri-DPO \citep{DBLP:conf/emnlp/PattnaikMOYM24} begins with examples showing large preference gaps. Other domain-specific efforts exist \citep{DBLP:conf/aaai/ZhaoWH21, upadhyay2025synlexlmscalinglegalllms, DBLP:conf/iclr/QiLILSSYYY00D25}. However, most rely on heuristics and expert knowledge. Our work aims to automate curriculum discovery via LLM agents.

\textbf{Model Merging.}
Model merging combines parameters from multiple models via arithmetic operations. \citet{DBLP:conf/icml/WortsmanIGRLMNF22} and Task Arithmetic \citep{DBLP:conf/iclr/IlharcoRWSHF23} show that adding or subtracting parameters can enhance robustness or transfer task skills. Techniques like TIES-Merging \citep{DBLP:conf/nips/YadavTCRB23}, DARE \citep{DBLP:conf/icml/Yu0Y0L24}, and many others \citep{DBLP:conf/acl/HuangLHCLHTL24, DBLP:conf/eccv/JangYH24, DBLP:conf/eccv/JangHSCL24, DBLP:journals/corr/abs-2412-04144, DBLP:conf/nips/Ortiz-JimenezFF23, DBLP:conf/iclr/LiuGS24} continue to expand the field. MergeKit \citep{goddard-etal-2024-arcees} facilitates implementation of merging techniques. Evolutionary methods \citep{DBLP:journals/natmi/AkibaSTSH25} and skill-efficient merging \citep{DBLP:conf/emnlp/MorrisonSHKDD24, DBLP:journals/corr/abs-2410-14735} optimize model merging parameters on taget tasks. Since merging and training are not independent, optimizing both jointly is crucial. This paper is the first to propose a unified approach that automates both training and merging through LLM agents to construct optimal pipelines.
\section{Conclusion}
In this work, we propose LaMDAgent, an automated framework for constructing post-training pipelines via LLM-based agents. Empirical results across two experimental settings demonstrate that LaMDAgent substantially outperforms all  baselines by autonomously identifying effective yet often-overlooked strategies by practitioners. To reduce the computational cost of pipeline exploration, we investigated scaling strategies and found that data-size scaling offers  benefits, whereas model-size scaling poses nontrivial challenges. 
These findings position LaMDAgent as a promising direction toward automating and systematizing post-training pipeline design, thereby reducing reliance on domain expertise and facilitating broader accessibility in LLM adaptation.
\section{Limitations}
Our experiments were conducted using Gemma 2 as the base model. It remains to be investigated how the outcomes might change when different base models are used. Furthermore, we only used English-language datasets. While our method is not expected to be highly language-dependent, it remains unclear whether it performs adequately on minority or low-resource languages.

In principle, the proposed framework allows for arbitrary action types. However, in this study, we focused on TIES-Merging and Supervised Fine-Tuning. It would be highly interesting to explore what kinds of pipelines could be discovered by combining our method with other model merging techniques, preference learning approaches, or data generation strategies.

Our experiments did not yield positive results in the context of model size scaling. Therefore, achieving positive transfer at larger scales may require further innovation in future work.

\bibliography{arxiv}

\appendix



\section{Templates}\label{appendix:templates}
Table~\ref{tab:action type selection template}, \ref{tab:objects selection template}, and \ref{tab:eval template} are prompt templates for action type selection, object selection, and memory generation in our proposed LaMDAgent, respectively. 

Table~\ref{tab:config} shows an example of configs for LaMDAgent.
\begin{figure*}[thb]
  \centering
\vspace{-10pt}
\begin{mybox}[Prompt template to select an action type]
\vspace{-9pt}
\lstset{basicstyle=\scriptsize\ttfamily}
\begin{lstlisting}
You are a developer of Large Language Models (LLMs) who tests model improvement strategies based on a given hypothesis. You are provided with Self-Reflections obtained from analyzing the result of a previous trial conducted for model improvement. Based on the Self-Reflections, select one action type from the Action Type List to create a more performant model. Analyze the Self-Reflections to identify the most promising action type, and provide the number of the selected action type at the end. If the Self-Reflections are not provided, please select randomly.

Self-Reflections:
<reflection>

Action List:
<action_types>

Selected Action Type NUMBER:
\end{lstlisting}
\vspace{-6pt}
\end{mybox}
\vspace{-9pt}
    \caption{Prompt template to select an action type.}
  \label{tab:action type selection template}
\end{figure*}

\begin{figure*}[thb]
  \centering
\vspace{-10pt}
\begin{mybox}[Prompt template to select objects]
\vspace{-9pt}
\lstset{basicstyle=\scriptsize\ttfamily}
\begin{lstlisting}
You are a developer of Large Language Models (LLMs). Your task is to determine a configuration for creating an LLM.
The configuration consists of multiple object types, and for each object type, you must select one object from a set of candidate objects.
To aid in your selection, you are provided with introspective analysis based on past LLM configurations and their outcomes.
Please output the selected objects in the order of the object types displayed, using comma separation and enclosed in [[ ]], e.g., [[1, 0, 2]] at the end of the output. If the Self-Reflections are not provided, please select randomly and think of a combination that has not been tried in the past trials. Also, the k-th model at step n is named in the format 0--n--k. Since such models also have promising potential, please include them in the search scope.

Self-Reflections:
<reflection>

Object Candidates:
<object_cands>

Selected Object NUMBERs:
\end{lstlisting}
\vspace{-6pt}
\end{mybox}
\vspace{-9pt}
    \caption{Prompt template to select objects.}
  \label{tab:objects selection template}
\end{figure*}

\begin{figure*}[thb]
  \centering
\vspace{-10pt}
\begin{mybox}[Prompt template to update memory]
\vspace{-9pt}
\lstset{basicstyle=\scriptsize\ttfamily}
\begin{lstlisting}
You are a developer of Large Language Models (LLMs) that can improve models based on self reflections. You will be given results and memories of the previous improving trials. The results consist of actions and scores, where the scores are out of 1 point. And also, You will be provided with newly aquired trials. In a few sentences, update your memories based on the previous trials, memoeries, and new results.  

# Previous Results
<previous results>

# Previous Memories Aquired from Previous Trials
<previous memories>

# Newly aquired Results
<new results>

Updated Memory:
\end{lstlisting}
\vspace{-6pt}
\end{mybox}
\vspace{-9pt}
    \caption{Prompt template to update memory.}
  \label{tab:eval template}
\end{figure*}

\begin{figure*}[thb]
  \centering
\vspace{-10pt}
\begin{mybox}[An example config of our proposed method]
\vspace{-9pt}
\lstset{basicstyle=\scriptsize\ttfamily}
\begin{lstlisting}
{
    "seed": 42,
    "total_timesteps": 100,
    "controller": "LaMDAgent_gpt",
    "controller_model": "gpt-4o-2024-11-20",
    "objects": {
        "base_models": ["models/gemma-2-2b"],
        "models": ["models/gemma-2-2b--gsm8k_1k", "models/gemma-2-2b--commonsense_qa_1k", "models/gemma-2-2b--trivia_qa_1k_w_context","models/gemma-2-2b"],
        "sft_dataset": ["data/sft_formatted/gsm8k_1k", "data/sft_formatted/commonsense_qa_1k", "data/sft_formatted/trivia_qa_1k_w_context", "data/sft_formatted/gsm1k_cqa1k_tqa1k"],
        "sft_lr": [0.000001],
        "ties_weights": [[0.5, 0.5]],
        "ties_density": [0.5],
    },    
    "action_types": {
        "sft": ["models", "sft_dataset", "sft_lr"],
        "ties_merging": ["base_models", "models", "models", "ties_weights", "ties_density"]                
    },
    "eval_tasks": [["gsm8k", "acc"], ["commonsenseqa", "acc"], ["trivia_qa_w_context", "acc"]],
    "score_aggregation": "mean"
}
\end{lstlisting}
\vspace{-6pt}
\end{mybox}
\vspace{-9pt}
    \caption{An example config of our proposed method.}
  \label{tab:config}
\end{figure*}


\end{document}